\documentclass[10pt,twocolumn,letterpaper]{article}

\usepackage{iccv}
\usepackage{times}
\usepackage{epsfig}
\usepackage{graphicx}
\usepackage{amsmath}
\usepackage{amssymb}
\usepackage[font=small]{caption}
\usepackage[hang,flushmargin]{footmisc}
\usepackage{pifont}
\usepackage{multirow}
\newcommand{\rf}[1]{{\textbf{\color{red}{#1}}}} 
\newcommand{\bd}[1]{{\color{blue}{#1}}} 
\newcommand{\cmark}{\ding{51}}%
\usepackage[pagebackref=true,breaklinks=true,letterpaper=true,colorlinks,bookmarks=false]{hyperref}

\iccvfinalcopy 

\ificcvfinal\fi

\begin{document}

\title{BasicVSR++: Improving Video Super-Resolution\\with Enhanced Propagation and Alignment}

\author{Kelvin C.K. Chan\qquad Shangchen Zhou\qquad  Xiangyu Xu\qquad Chen Change Loy\textsuperscript{*}\\
S-Lab, Nanyang Technological University\\
{\tt\small \{chan0899, s200094, xiangyu.xu, ccloy\}@ntu.edu.sg}
}

\ificcvfinal\thispagestyle{empty}\fi
\thispagestyle{empty}
\twocolumn[{%
            \renewcommand\twocolumn[1][]{#1}%
            \vspace{-1em}
            \maketitle
            \vspace{-1em}
            \begin{center}
                \centering
                \vspace{-0.5cm}
                \includegraphics[width=0.99\textwidth]{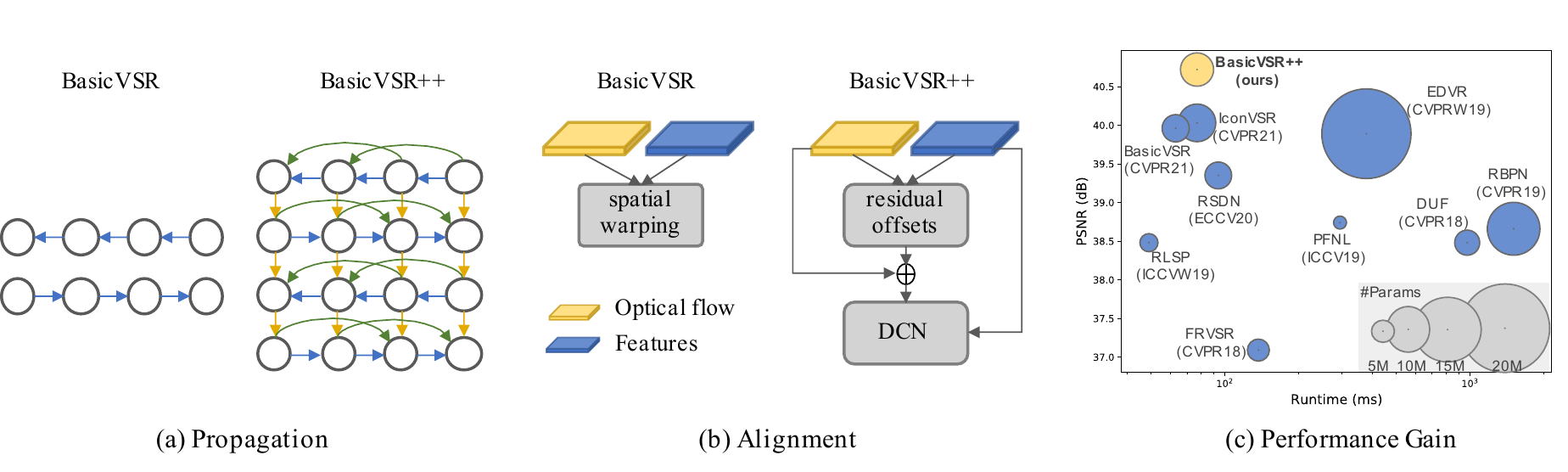}\vspace{0.1cm}
                \vskip -0.2cm
                \captionof{figure}{\textbf{Improvements over BasicVSR~\cite{chan2021basicvsr}.} \textbf{(a)} Second-order grid propagation in BasicVSR++ allows a more effective propagation of features. \textbf{(b)} Flow-guided deformable alignment in BasicVSR++ provides a means for more robust feature alignment across misaligned frames. \textbf{(c)} BasicVSR++ outperforms existing state of the arts while maintaining efficiency.}
                \label{fig:teaser}
                \vspace{0.15cm}
            \end{center}%
        }]

\begin{abstract}
    \vspace{-0.1cm}
    A recurrent structure is a popular framework choice for the task of video super-resolution. The state-of-the-art method BasicVSR adopts bidirectional propagation with feature alignment to effectively exploit information from the entire input video.
    \makeatletter{\renewcommand*{\@makefnmark}{}
        \footnotetext{$^*$Corresponding author}\makeatother}
    In this study, we redesign BasicVSR by proposing second-order grid propagation and flow-guided deformable alignment. We show that by empowering the recurrent framework with the enhanced propagation and alignment, one can exploit spatiotemporal information across misaligned video frames more effectively.
    The new components lead to an improved performance under a similar computational constraint.
    In particular, our model BasicVSR++ surpasses BasicVSR by 0.82~dB in PSNR with similar number of parameters. In addition to video super-resolution, BasicVSR++ generalizes well to other video restoration tasks such as compressed video enhancement.
    In NTIRE 2021, BasicVSR++ obtains three champions and one runner-up in the Video Super-Resolution and Compressed Video Enhancement Challenges. Codes and models will be released to MMEditing\footnote{\url{https://github.com/open-mmlab/mmediting}}.
\end{abstract}

\section{Introduction}
Video super-resolution (VSR) is challenging in that one needs to gather complementary information across misaligned video frames for restoration. One prevalent approach is the sliding-window framework~\cite{haris2019recurrent,tian2020tdan,wang2019edvr,xue2019video}, where each frame in the video is restored using the frames within a short temporal window.
In contrast to the sliding-window framework, a recurrent framework attempts to exploit the long-term dependencies by propagating the latent features. In general, these methods~\cite{fuoli2019efficient,huang2015bidirectional,huang2018video,isobe2020video1,isobe2020revisiting, sajjadi2018frame} allow a more compact model compared to those in the sliding-window framework. Nevertheless, the problems of transmitting long-term information and aligning features across frames in a recurrent model remain formidable.

A recent work by Chan~\etal~\cite{chan2021basicvsr} studies the problems carefully. It summarizes the common VSR pipelines into four components, namely \textit{Propagation}, \textit{Alignment}, \textit{Aggregation}, and \textit{Upsampling}, and proposes BasicVSR. In BasicVSR, bidirectional propagation is adopted to exploit information from the entire input video for reconstruction. For alignment, optical flow is adopted for feature warping. BasicVSR serves as a succinct yet strong backbone where components can be easily added for performance gain.
However, its rudimentary designs in propagation and alignment limit the efficacy of information aggregation. As a result, the network often struggles to restore fine details, especially when dealing with occluded and complex regions. The shortcomings call for refined designs in propagation and alignment.

In this work, we redesign BasicVSR by devising \textit{second-order grid propagation} and \textit{flow-guided deformable alignment} that allow information to be propagated and aggregated more effectively:

\noindent
1) The proposed second-order grid propagation, as shown in Fig.~\ref{fig:teaser}(a), addresses two limitations in BasicVSR: i) we allow more aggressive bidirectional propagation arranged in a grid-like manner, and ii) we relax the assumption of first-order Markov property in BasicVSR, and incorporate a second-order connection~\cite{soltani2016higher} into the network so that information can be aggregated from different spatiotemporal locations. Both modifications ameliorate information flow in the network and improve robustness of the network against occluded and fine regions.

\noindent
2) BasicVSR shows advantages of using optical flow for temporal alignment. However, optical flow is not robust to occlusion. Inaccurate flow estimation could jeopardize the restoration performance. Deformable alignment~\cite{tian2020tdan,wang2019deformable,wang2019edvr} has demonstrated its superiority in VSR, but it is difficult to train in practice~\cite{chan2021understanding}. To take advantage of deformable alignment while overcoming the training instability, we propose flow-guided deformable alignment, as shown in Fig.~\ref{fig:teaser}(b). In the proposed module, instead of learning the DCN offsets directly~\cite{dai2017deformable,zhu2018deformable}, we reduce the burden of offset learning by using optical flow field as base offsets refined by flow field residue. The latter can be learned more stably than the original DCN offsets.

The two aforementioned components are novel and more discussion can be found in the related work section.
Benefit from the more effective designs, BasicVSR++ can adopt a more lightweight backbone than its counterparts. Consequently, BasicVSR++ surpasses existing state of the arts, including BasicVSR and IconVSR (the more elaborated BasicVSR variant), by a large margin while maintaining efficiency (Fig.~\ref{fig:teaser}(c)). In particular, when compared to its precedent BasicVSR, a gain of 0.82~dB in PSNR on REDS4~\cite{wang2019edvr} is obtained with similar numbers of parameters. In addition, BasicVSR++ obtains three champions and one runner-up in the NTIRE 2021 Video Super-Resolution~\cite{son2021ntire} and Compressed Video Enhancement~\cite{yang2021ntire} Challenges.

\section{Related Work}
\noindent\textbf{Recurrent Networks.}
The recurrent framework is a popular structure adopted in various video processing tasks such as super-resolution~\cite{fuoli2019efficient,huang2015bidirectional,huang2018video,isobe2020video1,isobe2020revisiting,sajjadi2018frame}, deblurring~\cite{nah2019recurrent,zhou2019spatiotemporal}, and frame interpolation~\cite{xiang2020zooming}.
For instance, RSDN~\cite{isobe2020video1} adopts unidirectional propagation with a recurrent detail structural block and a hidden state adaptation module to enhance the robustness to appearance change and error accumulation.
Chan~\etal~\cite{chan2021basicvsr} propose BasicVSR. The work demonstrates the importance of bidirectional propagation over unidirectional propagation to better exploit features temporally. In addition, the study also shows the advantage of feature alignment in aligning highly relevant but misaligned features. We refer readers to~\cite{chan2021basicvsr} for the detailed comparisons of these components against the more conventional ways of performing propagation and alignment. In our experiments, we focus on comparing with BasicVSR since it is the state-of-the-art method for VSR.

\begin{figure*}[!t]
    \begin{center}
        \includegraphics[width=0.99\textwidth]{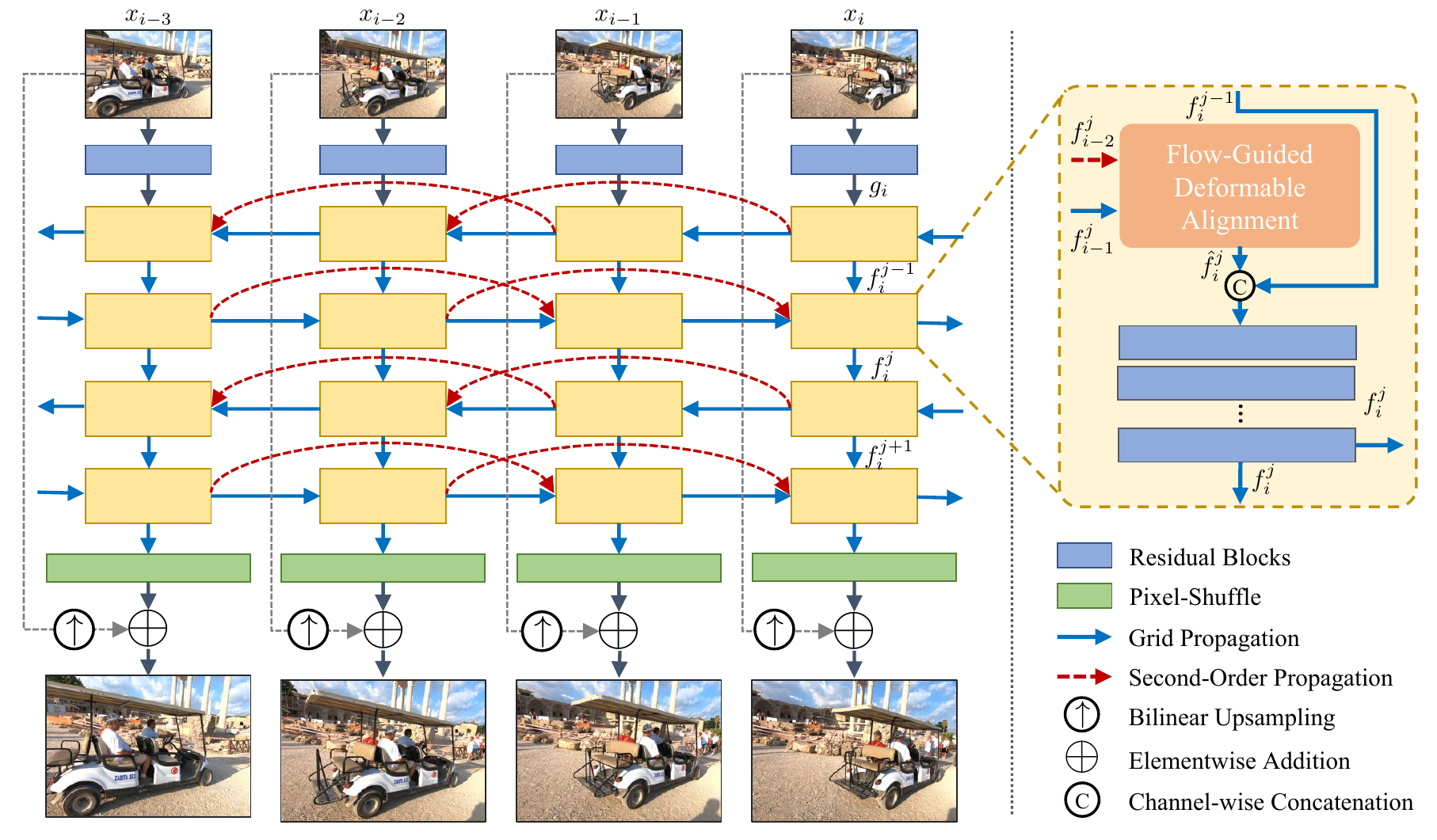}\vspace{-0.2cm}
        \caption{\textbf{An Overview of BasicVSR++.} BasicVSR++ consists of two modifications to improve propagation and alignment. For propagation, we introduce second-order propagation (blue solid lines) to refine features bidirectionally. In addition, second-order connection (red dotted lines) is adopted to improve the robustness of propagation. Within each propagation branch, flow-guided deformable alignment is proposed to increase the offset diversity while overcoming the offset overflow problem.}
        \label{fig:overview}
    \end{center}
    \vspace{-0.5cm}
\end{figure*}

\noindent\textbf{Grid Connections.}
Grid-like designs are seen in various vision tasks such as object detection~\cite{chen2018optimizing,sun2019high,wang2020deep}, semantic segmentation~\cite{fourure2017residual,sun2019high,wang2020deep,zhuang2019shelf}, and frame interpolation~\cite{niklaus2020softmax}. In general, these designs decompose a given image/feature into multiple resolutions, and grids are adopted \textit{across resolutions} to capture both fine and coarse information.
Unlike aforementioned methods, BasicVSR++ does not adopt a multi-scale design. Instead, the grid structure is designed for propagation \textit{across time} in a bidirectional fashion. We link different frames with a grid connection to repeatedly refine the features, improving expressiveness.

\noindent\textbf{Higher-Order Propagation.}
Higher-order propagation has been studied to improve gradient flow~\cite{ke2018sparse,lin1996learning,soltani2016higher}. These methods demonstrate improvements in different tasks including classification~\cite{ke2018sparse} and language modeling~\cite{soltani2016higher}. However, these methods do not consider temporal alignment, which is shown critical in the task of VSR~\cite{chan2021basicvsr}. To allow temporal alignment in second-order propagation, we incorporate alignment into our propagation scheme by extending our flow-guided deformable alignment to the second-order settings.

\noindent\textbf{Deformable Alignment.} Several works~\cite{tian2020tdan,wang2019deformable,wang2019edvr,xu2020learning} employ deformable alignment. TDAN~\cite{tian2020tdan} performs alignment at the feature level using deformable convolution. EDVR~\cite{wang2019edvr} further proposes a Pyramid Cascading Deformable (PCD) alignment with a multi-scale design. Recently, Chan~\etal~\cite{chan2021understanding} analyze deformable alignment and show that the performance gain over flow-based alignment comes from the offset diversity.
Motivated by~\cite{chan2021understanding}, we adopt deformable alignment but with a reformulation to overcome the training instability~\cite{chan2021understanding}.
Our flow-guided deformable alignment is different from offset-fidelity loss~\cite{chan2021understanding}. The latter uses optical flow as a loss function during training. In contrast, we directly incorporate optical flow into our module as base offsets, allowing a more explicit guidance, both during training and inference.

\section{Methodology}

BasicVSR++ consists of two effective modifications for improving \textit{propagation} and \textit{alignment}. As shown in Fig.~\ref{fig:overview}, given an input video, residual blocks are first applied to extract features from each frame. The features are then propagated under our second-order grid propagation scheme, where alignment is performed by our flow-guided deformable alignment. After propagation, the aggregated features are used to generate the output image through convolution and pixel-shuffling.

\subsection{Second-Order Grid Propagation}
\label{sec:propagation}
Most existing methods adopt unidirectional propagation~\cite{isobe2020video1,isobe2020revisiting,sajjadi2018frame}. Several works~\cite{chan2021basicvsr,huang2015bidirectional,huang2018video} adopt bidirectional propagation for exploiting the information available in the video sequence.
In particular, IconVSR~\cite{chan2021basicvsr} consists of a coupled propagation scheme with sequentially-connected branches to facilitate information exchange.

Motivated by the effectiveness of the bidirectional propagation, we devise a grid propagation scheme to enable \textit{repeated refinement through propagation}. More specifically, the intermediate features are propagated backward and forward in time in an alternating manner. Through propagation, the information from different frames can be ``revisited'' and adopted for feature refinement. Compared to existing works that propagate features only once, grid propagation repeatedly extracts information from the entire sequence, improving feature expressiveness.

To further enhance the robustness of propagation, we relax the assumption of first-order Markov property in BasicVSR and adopt a second-order connection, realizing a second-order Markov chain. With this relaxation, information can be aggregated from different spatiotemporal locations, improving robustness and effectiveness in occluded and fine regions.

Integrating the above two components, we devise our second-order grid propagation as follows. Let $x_i$ be the input image, $g_i$ be the feature extracted from $x_i$ by multiple residual blocks, and $f_i^j$ be the feature computed at the $i$-th timestep in the $j$-th propagation branch. In this section, we describe the procedure for forward propagation, and the procedure for backward propagation is defined similarly.

To compute the feature $f_i^j$, we first align $f_{i-1}^j$ and $f_{i-2}^j$ (following the second-order Markov chain) using our proposed flow-guided deformable alignment, which will be discussed in the next section:
\begin{equation}
    \hat{f}_i^j = \mathcal{A}\left(g_i, f_{i-1}^j, f_{i-2}^j, s_{i\rightarrow i-1}, s_{i\rightarrow i-2}\right),
\end{equation}
where $s_{i\rightarrow i-1}, s_{i\rightarrow i-2}$ denote the optical flows from $i$-th frame to the $(i{-}1)$-th and $(i{-}2)$-th frames, respectively, and $\mathcal{A}$ represents flow-guided deformable alignment\footnote{$s_{0\rightarrow-1}{=}s_{0\rightarrow-2}{=}s_{1\rightarrow-1}{=}f_{-1}{=}f_{-2}{=}0$.}. The features are then concatenated and passed into a stack of residual blocks:
\begin{equation}
    f_i^j = \hat{f}_i^j + \mathcal{R}\left(c\left(f_i^{j-1}, \hat{f}_{i}^j\right)\right),
\end{equation}
where $f^{0}_i = g_i$, $\mathcal{R}$ denotes the residual blocks, and $c$ denotes concatenation along channel dimension.

\begin{figure}[t]
    \begin{center}
        \includegraphics[width=0.45\textwidth]{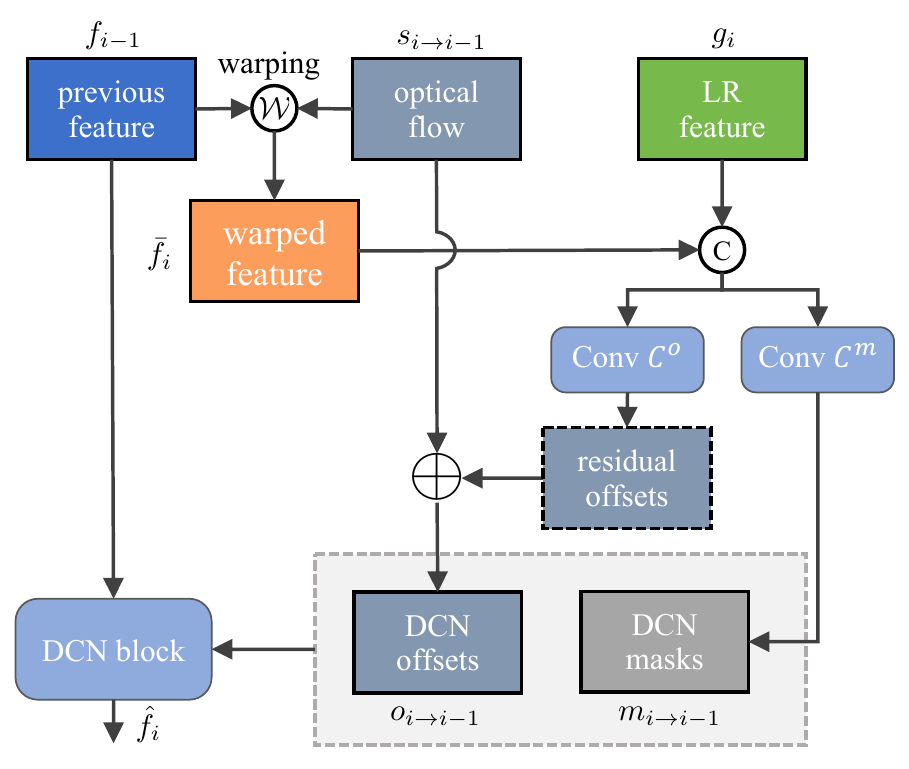}
        \caption{\textbf{Flow-guided deformable alignment.} Optical flow is used to pre-align the features. The aligned features are then concatenated to produce to DCN offsets (residue to optical flow). A DCN is then applied to the unwarped features. Only first-order connections are drawn, the second-order connections are omitted for simplicity.}
        \label{fig:flow-guided}
    \end{center}
    \vspace{-0.5cm}
\end{figure}

\subsection{Flow-Guided Deformable Alignment}
\label{sec:alignment}
Deformable alignment~\cite{wang2019deformable,wang2019edvr} has demonstrated significant improvements over flow-based alignment~\cite{haris2019recurrent,xue2019video} thanks to the offset diversity~\cite{chan2021understanding} intrinsically introduced in deformable convolution (DCN)~\cite{dai2017deformable,zhu2018deformable}.
However, deformable alignment module can be difficult to train~\cite{chan2021understanding}. The training instability often results in offset overflow, deteriorating the final performance.

To take advantage of the offset diversity while overcoming the instability, we propose to employ optical flow to guide deformable alignment, motivated by the strong relation between deformable alignment and flow-based alignment~\cite{chan2021understanding}. The graphical illustration is shown in Fig.~\ref{fig:flow-guided}. In the rest of this section, we detail the alignment procedure for forward propagation. The procedure for backward propagation is defined similarly. The superscript $j$ is omitted for notational simplicity.

At the $i$-th timestep, given the feature $g_i$ computed from the $i$-th LR image, the feature $f_{i-1}$ computed for the previous timestep, and the optical flow $s_{i\rightarrow i-1}$ to the previous frame, we first warp $f_{i-1}$ with $s_{i\rightarrow i-1}$:
\begin{equation}
    \bar{f}_{i-1} = \mathcal{W}(f_{i-1}, s_{i\rightarrow i-1}),
\end{equation}
where $\mathcal{W}$ denotes the spatial warping operation.
The pre-aligned features are then used to compute the DCN offsets $o_{i\rightarrow i-1}$ and modulation masks $m_{i\rightarrow i-1}$. Instead of directly computing the DCN offsets, we compute the residue to the optical flow:
\begin{equation}
    \begin{split}
        &o_{i\rightarrow i-1} = s_{i\rightarrow i-1} + \mathcal{C}^o\left(c(g_i, \bar{f}_{i-1})\right),\\
        &m_{i\rightarrow i-1} = \sigma\left(\mathcal{C}^m\left(c(g_i, \bar{f}_{i-1})\right)\right).
    \end{split}
\end{equation}
Here $\mathcal{C}^{\{o, m\}}$ denotes a stack of convolutions, and $\sigma$ denotes the sigmoid function.
A DCN is then applied to the unwarped feature $f_{i-1}$:
\begin{equation}
    \hat{f}_i = \mathcal{D}\left(f_{i-1}; o_{i\rightarrow i-1}, m_{i\rightarrow i-1}\right),
\end{equation}
where $\mathcal{D}$ denotes a deformable convolution.

The above formulation is designed only for aligning one single feature, and hence is not directly applicable to our second-order propagation. The most intuitive way to adapt to the second-order settings is to apply the above procedure to the two features, $f_{i-1}^j$ and $f_{i-2}^j$, independently. However, this requires doubled computations, resulting in reduced efficiency. Furthermore, separate alignment potentially ignores the complementary information from the features. Therefore, we allow alignment of two features simultaneously. More specifically, we concatenate the warped features and flows to compute the offsets $o_{i-p}$ ($p{=}1, 2$):
\begin{equation}
    \label{eq:eq}
    \begin{split}
        &o_{i\rightarrow i-p} = s_{i\rightarrow i-p} + \mathcal{C}^o\left(c(g_i, \bar{f}_{i-1}, \bar{f}_{i-2})\right),\\
        &m_{i\rightarrow i-p} = \sigma\left(\mathcal{C}^m\left(c(g_i, \bar{f}_{i-1}, \bar{f}_{i-2})\right)\right).
    \end{split}
\end{equation}
A DCN is then applied to the unwarped features:
\begin{equation}
    \begin{split}
        &o_i = c(o_{i\rightarrow i-1}, o_{i\rightarrow i-2}),\\
        &m_i = c(m_{i\rightarrow i-1},m_{i\rightarrow i-2}),\\
        &\hat{f}_i = \mathcal{D}\left(c(f_{i-1}, f_{i-2}); o_i, m_i \right).
    \end{split}
\end{equation}
More details of the second-order flow-guided deformable alignment are provided in the supplementary material.
\begin{table*}[!t]
    \caption{\textbf{Quantitative comparison (PSNR/SSIM).} All results are calculated on Y-channel except REDS4~\cite{nah2019ntire} (RGB-channel). \rf{Red} and \bd{blue} colors indicate the best and the second-best performance, respectively. The runtime is computed on an LR size of $180{\times}320$. A $4\times$ upsampling is performed following previous studies. Blanked entries correspond to results not reported in previous works.}
    \label{tab:quan}
    \vspace{-0.4cm}
    \begin{center}\scalebox{0.86}{
            \tabcolsep=0.1cm
            \begin{tabular}{l|c|c||c|c|c||c|c|c}
                \hline
                \multirow{2}{*}{}                       &            &              & \multicolumn{3}{c||}{BI degradation} & \multicolumn{3}{c}{BD degradation}                                                                                                                                \\ \cline{2-9}
                                                        & Params (M) & Runtime (ms) & REDS4~\cite{nah2019ntire}            & Vimeo-90K-T~\cite{xue2019video}    & Vid4~\cite{liu2014bayesian} & UDM10~\cite{yi2019progressive} & Vimeo-90K-T~\cite{xue2019video} & Vid4~\cite{liu2014bayesian} \\ \hline
                Bicubic                                 & -          & -            & 26.14/0.7292                         & 31.32/0.8684                       & 23.78/0.6347                & 28.47/0.8253                   & 31.30/0.8687                    & 21.80/0.5246                \\
                VESPCN~\cite{caballero2017real}         & -          & -            & -                                    & -                                  & 25.35/0.7557                & -                              & -                               & -                           \\
                SPMC~\cite{tao2017detail}               & -          & -            & -                                    & -                                  & 25.88/0.7752                & -                              & -                               & -                           \\
                TOFlow~\cite{xue2019video}              & -          & -            & 27.98/0.7990                         & 33.08/0.9054                       & 25.89/0.7651                & 36.26/0.9438                   & 34.62/0.9212                    & -                           \\
                FRVSR~\cite{sajjadi2018frame}           & 5.1        & 137          & -                                    & -                                  & -                           & 37.09/0.9522                   & 35.64/0.9319                    & 26.69/0.8103                \\
                DUF~\cite{jo2018deep}                   & 5.8        & 974          & 28.63/0.8251                         & -                                  & -                           & 38.48/0.9605                   & 36.87/0.9447                    & 27.38/0.8329                \\
                RBPN~\cite{haris2019recurrent}          & 12.2       & 1507         & 30.09/0.8590                         & 37.07/0.9435                       & 27.12/0.8180                & 38.66/0.9596                   & 37.20/0.9458                    & -                           \\
                EDVR-M~\cite{wang2019edvr}              & 3.3        & 118          & 30.53/0.8699                         & 37.09/0.9446                       & 27.10/0.8186                & 39.40/0.9663                   & 37.33/0.9484                    & 27.45/0.8406                \\
                EDVR~\cite{wang2019edvr}                & 20.6       & 378          & 31.09/0.8800                         & \bd{37.61}/\bd{0.9489}             & 27.35/0.8264                & 39.89/0.9686                   & 37.81/0.9523                    & 27.85/0.8503                \\
                PFNL~\cite{yi2019progressive}           & 3.0        & 295          & 29.63/0.8502                         & 36.14/0.9363                       & 26.73/0.8029                & 38.74/0.9627                   & -                               & 27.16/0.8355                \\
                MuCAN~\cite{li2020mucan}                & -          & -            & 30.88/0.8750                         & 37.32/0.9465                       & -                           & -                              & -                               & -                           \\
                TGA~\cite{isobe2020video}               & 5.8        & -            & -                                    & -                                  & -                           & -                              & 37.59/0.9516                    & 27.63/0.8423                \\
                RLSP~\cite{fuoli2019efficient}          & 4.2        & 49           & -                                    & -                                  & -                           & 38.48/0.9606                   & 36.49/0.9403                    & 27.48/0.8388                \\
                RSDN~\cite{isobe2020video1}             & 6.2        & 94           & -                                    & -                                  & -                           & 39.35/0.9653                   & 37.23/0.9471                    & 27.92/0.8505                \\
                RRN~\cite{isobe2020revisiting}          & 3.4        & 45           & -                                    & -                                  & -                           & 38.96/0.9644                   & -                               & 27.69/0.8488                \\
                \mbox{BasicVSR}~\cite{chan2021basicvsr} & 6.3        & 63           & 31.42/0.8909                         & 37.18/0.9450                       & 27.24/0.8251                & 39.96/\bd{0.9694}              & 37.53/0.9498                    & 27.96/0.8553                \\
                \mbox{IconVSR}~\cite{chan2021basicvsr}  & 8.7        & 70           & \bd{31.67}/\bd{0.8948}               & 37.47/0.9476                       & \bd{27.39}/\bd{0.8279}      & \bd{40.03}/\bd{0.9694}         & \bd{37.84}/\bd{0.9524}          & \bd{28.04}/\bd{0.8570}      \\ \hline
                \textbf{BasicVSR++}                     & 7.3        & 77           & \rf{32.39}/\rf{0.9069}               & \rf{37.79}/\rf{0.9500}             & \rf{27.79}/\rf{0.8400}      & \rf{40.72}/\rf{0.9722}         & \rf{38.21}/\rf{0.9550}          & \rf{29.04}/\rf{0.8753}      \\\hline
            \end{tabular}}
        \vspace{-0.5cm}
    \end{center}
\end{table*}

\noindent\textbf{Discussion.} Unlike existing methods~\cite{tian2020tdan,wang2019deformable,wang2019edvr,xu2020learning} that directly compute the DCN offsets, our proposed flow-guided deformable alignment adopts optical flow as guidance. The benefits are two-fold. First, since CNNs are known to have local receptive fields, the learning of offsets can be assisted by pre-aligning the features using optical flow.
Second, by learning only the residue, the network needs to learn only small deviations from the optical flow, reducing the burden in typical deformable alignment modules.
In addition, instead of directly concatenating the warped feature, the modulation masks in DCN act as attention maps to weigh the contributions of different pixels, providing additional flexibility.
\begin{table}[!t]
    \caption{\textbf{Performance of a lighter BasicVSR++.} Our lighter model, BasicVSR++ (S), has a similar complexity to BasicVSR and IconVSR, but still shows considerable improvements. The PSNR and runtime are computed on REDS4.}
    \vspace{-0.4cm}
    \label{tab:small_model}
    \begin{center}
        \tabcolsep=0.13cm
        \scalebox{0.9}{
            \begin{tabular}{l|c|c||c}
                \hline
                             & BasicVSR~\cite{chan2021basicvsr} & IconVSR~\cite{chan2021basicvsr} & \textbf{BasicVSR++ (S)} \\\hline
                Params (M)   & 6.3                              & 8.7                             & 6.4                     \\
                Runtime (ms) & 63                               & 70                              & 69                      \\
                PSNR (dB)    & 31.42                            & 31.67                           & 32.24                   \\\hline
            \end{tabular}}
    \end{center}
    \vspace{-0.5cm}
\end{table}

\section{Experiments}
\label{sec:exp}
Two widely-used datasets are adopted for training: REDS~\cite{nah2019ntire} and Vimeo-90K~\cite{xue2019video}.
For REDS, following BasicVSR~\cite{chan2021basicvsr}, we use REDS4\footnote{Clips 000, 011, 015, 020 of REDS training set.} as our test set and REDSval4\footnote{Clips 000, 001, 006, 017 of REDS validation set.} as our validation set. The remaining clips are used for training.
We use Vid4~\cite{liu2014bayesian}, UDM10~\cite{yi2019progressive}, and Vimeo-90K-T~\cite{xue2019video} as test sets along with Vimeo-90K.
All models are tested with $4{\times}$ downsampling using two degradations -- Bicubic (BI) and Blur Downsampling (BD).

We adopt Adam optimizer~\cite{kingma2014adam} and Cosine Annealing scheme~\cite{loshchilov2016sgdr}. The initial learning rate of the main network and the flow network are set to $1{\times}10^{-4}$ and $2.5{\times}10^{-5}$, respectively. The total number of iterations is 600K, and the weights of the flow network are fixed during the first 5,000 iterations. The batch size is 8 and the patch size of input LR frames is $64{\times}64$. We use Charbonnier loss~\cite{charbonnier1994two} since it better handles outliers and improves the performance over the conventional $\ell_2$-loss~\cite{lai2017deep}.
We use pre-trained SPyNet~\cite{ranjan2017optical} as our flow network. Its parameters and runtime are considered inclusively in our method. The number of residual blocks for each branch is set to $7$. The number of feature channels is $64$. Detailed experimental settings and model architectures are provided in the supplementary material.

\begin{figure*}[!t]
    \begin{center}
        \includegraphics[width=0.99\textwidth]{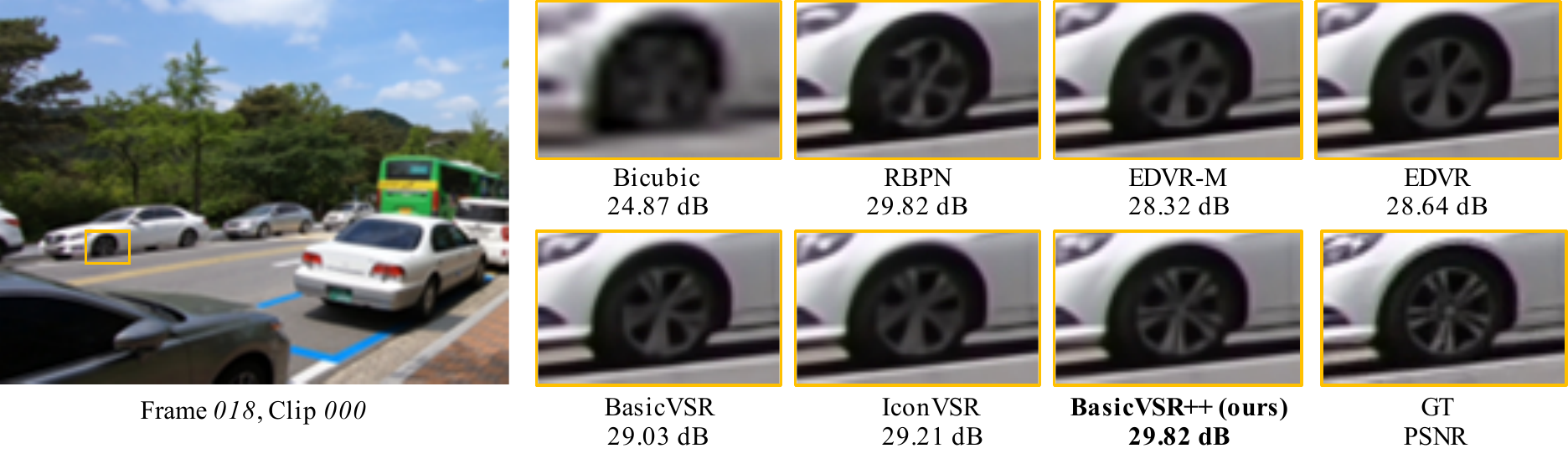}
        \vskip -0.35cm
        \caption{\textbf{Challenging scenario on REDS4~\cite{wang2019edvr}.} Only BasicVSR++ is able to recover the patterns of the wheel's spokes.}
        \label{fig:reds4}
    \end{center}
    \vspace{-0.2cm}
\end{figure*}
\begin{figure*}[!t]
    \begin{center}
        \includegraphics[width=0.99\textwidth]{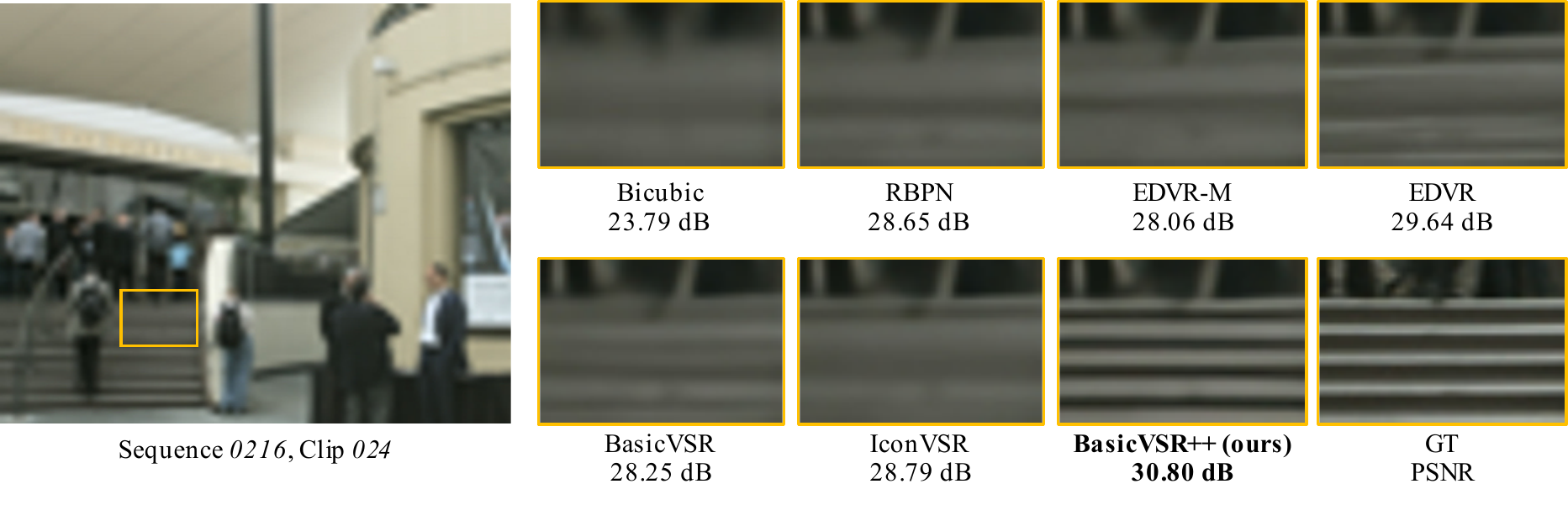}
        \vskip -0.35cm
        \caption{\textbf{Challenging scenario on Vimeo-90K-T~\cite{xue2019video}.} Only BasicVSR++ is able to reconstruct the stairs.}
        \label{fig:vimeo}
    \end{center}
    \vspace{-0.2cm}
\end{figure*}
\begin{figure*}[!t]
    \begin{center}
        \includegraphics[width=0.99\textwidth]{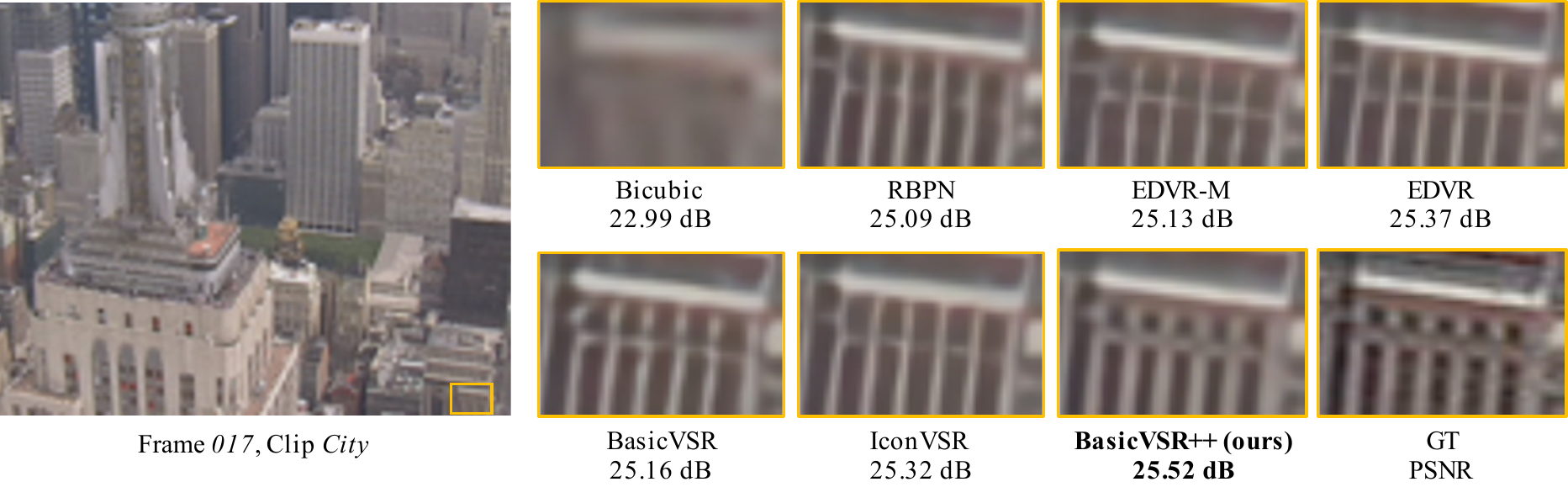}
        \vskip -0.35cm
        \caption{\textbf{Challenging scenario on Vid4~\cite{liu2014bayesian}.} Only BasicVSR++ is able to recover the correct structure of the building.}
        \label{fig:vid4}
    \end{center}
    \vspace{-0.35cm}
\end{figure*}
\subsection{Comparisons with State-of-the-Art Methods}
We conduct comprehensive experiments by comparing with 16 models, as listed in Table~\ref{tab:quan}.
The quantitative results are summarized in Table~\ref{tab:quan} and the speed and performance comparison is provided in Fig.~\ref{fig:teaser}(c). Note that the parameters reported above are inclusive of that in the optical flow network (if any). So the comparison is fair.

As shown in Table~\ref{tab:quan}, BasicVSR++ achieves state-of-the-art performance on all datasets for both degradations. In particular, BasicVSR++ outperforms EDVR~\cite{wang2019edvr}, a large-capacity sliding-window method, by up to 1.3~dB in PSNR, while having 65\% fewer parameters. When compared to the previous state of the art, IconVSR~\cite{chan2021basicvsr}, BasicVSR++ possesses fewer parameters but has improvements of up to 1~dB. As shown in Table~\ref{tab:small_model}, even if we train a lighter version of BasicVSR++ (denoted as BasicVSR++ (S)) with comparable network parameters and runtime to BasicVSR and IconVSR, our model still shows an improvement of 0.82~dB over BasicVSR and 0.57~dB over IconVSR. Such gains are considered significant in VSR.

Some qualitative comparisons are shown in Fig.~\ref{fig:reds4} to Fig.~\ref{fig:vid4}. BasicVSR++ successfully restores the fine details. In particular, BasicVSR++ is the only method that restores the wheel's spokes in Fig.~\ref{fig:reds4}, the stairs in Fig.~\ref{fig:vimeo}, and the building structure in Fig.~\ref{fig:vid4}. More examples are provided in the supplementary material.

\begin{figure*}[!t]
    \begin{center}
        \includegraphics[width=0.99\textwidth]{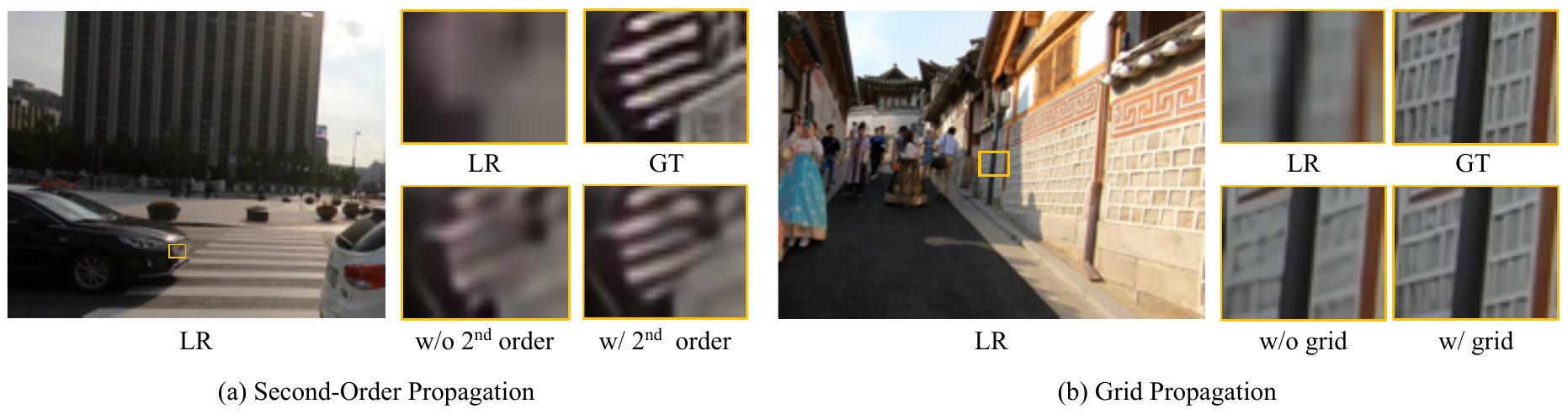}
        \vskip -0.3cm
        \caption{\textbf{Analysis of second-order grid propagation.} By propagating the features more effectively, our second-order grid propagation leads to more details, improving the output quality.}
        \label{fig:propagation}
    \end{center}
    \vspace{-0.5cm}
\end{figure*}
\begin{figure*}[!t]
    \begin{center}
        \includegraphics[width=0.99\textwidth]{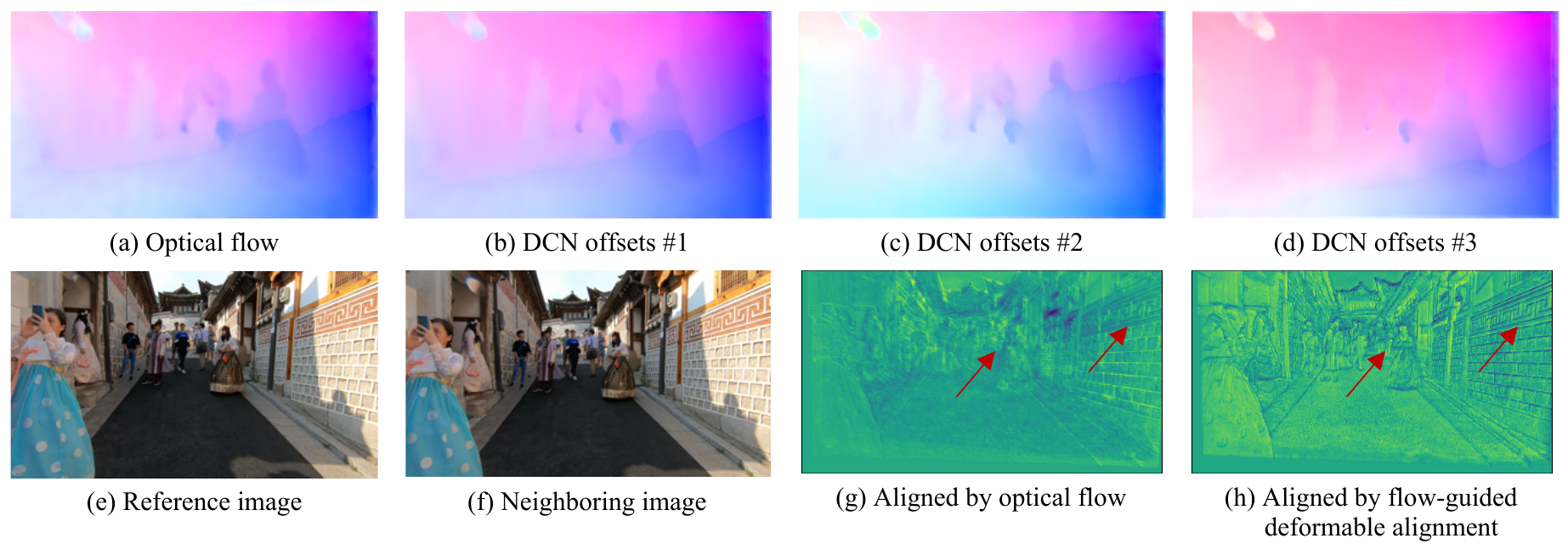}
        \vskip -0.3cm
        \caption{\textbf{Analysis of flow-guided deformable alignment.} \textbf{(a-d)} The DCN offsets are highly similar to optical flow, but still with noticeable differences. \textbf{(e-f)} The reference and neighboring images. \textbf{(g)} The feature aligned by optical flow experiences blurry edges. \textbf{(h)} The feature aligned by our proposed module is sharper and preserves more details, as indicated by the red arrows.}
        \label{fig:alignment}
    \end{center}
    \vspace{-0.5cm}
\end{figure*}

\section{Ablation Studies}
\label{sec:ablation}
To understand the contributions of the proposed components, we start with a baseline and gradually insert the components.
From Table~\ref{tab:ablation}, it is apparent that each component brings considerable improvement, ranging from 0.14~dB to 0.46~dB in PSNR.

In theory, our proposed propagation schemes can be extended to higher orders and more propagation iterations. However, while the performance gain is considerable when increasing from first-order to second-order (\ie~(B)$\rightarrow$(C)), and from one to two iterations (\ie~(C)$\rightarrow$BasicVSR++), we observe in our preliminary experiments that further increasing the orders and number of iterations does not lead to a significant improvement ($0.05$~dB in PSNR). Therefore, we keep both the orders and iterations to two.

\begin{table}[!t]
    \caption{\textbf{Ablation studies of the components.} Each component brings significant improvements in PSNR, verifying their effectiveness.}
    \vspace{-0.4cm}
    \label{tab:ablation}
    \begin{center}
        \tabcolsep=0.13cm
        \scalebox{0.85}{
            \begin{tabular}{l|c|c|c|c}
                \hline
                                           & (A)   & (B)    & (C)    & \textbf{BasicVSR++} \\\hline
                Flow-Guided Deform. Align. &       & \cmark & \cmark & \cmark              \\
                Second-Order Propagation   &       &        & \cmark & \cmark              \\
                Grid Propagation           &       &        &        & \cmark              \\\hline\hline
                PSNR (dB)                  & 31.48 & 31.94  & 32.08  & 32.39               \\\hline
            \end{tabular}}
    \end{center}
    \vspace{-0.7cm}
\end{table}

\noindent
\textbf{Second-Order Grid Propagation}.
We further provide some qualitative comparisons to understand the contributions of the proposed propagation scheme.
As shown in the two examples of Fig.~\ref{fig:propagation}, the contribution of both the second-order propagation and grid propagation is more noticeable in regions that contain fine details and complex textures.
In those regions, there is limited information from the current frame that can be employed for reconstruction. To improve the output quality of those regions, effective information aggregation from other video frames is necessary.
With our second-order propagation scheme, the information can be transmitted via a robust and effective propagation. This complementary information essentially assists the restoration of the fine details. As shown in the examples, the network successfully restores the details with our components, whereas the counterparts without our components produce blurry outputs.

\noindent
\textbf{Flow-Guided Deformable Alignment}.
In Fig.~\ref{fig:alignment}(a-d), we compare the offsets with the optical flow computed by the flow estimation module in BasicVSR++.
By learning only the residue to optical flow, the network produces offsets that are highly similar to the optical flow, but with observable differences. When compared to the baseline which aggregates information from only one spatial location indicated by the motion (optical flow), our proposed module allows retrieving information from multiple locations around, providing additional flexibility.

This flexibility leads to features with better quality, as shown in Fig.~\ref{fig:alignment}(g-h). When the warping is performed by using optical flow, the aligned features contain blurry edges, owing to the interpolation operation in spatial warping. In contrast, by gathering more information from the neighbors, the feature aligned by our proposed module is sharper and preserves more details.

To demonstrate the superiority of our designs, we compare our alignment module with two variants: (1) No optical flow is used. (2) Optical flow is used as in the offset-fidelity loss~\cite{chan2021understanding}, \ie the flow is merely used as supervision in the loss function (rather than serving as base offsets as in our method). As shown in Table~\ref{tab:alignment}, without using optical flow as guidance, the instability causes training to collapse, leading to a very poor PSNR value.
When using the offset-fidelity loss, the training is stabilized. However, a drop of 2.17~dB from our full model is observed. Our flow-guided deformable alignment directly incorporates optical flow into the network to provide more explicit guidance, leading to better results.
\begin{table}[t]
    \caption{\textbf{Comparison of alignment modules.} Using optical flow to guide deformable alignment successfully stabilizes training. BasicVSR++ directly incorporates optical flow into the network, outperforming the offset-fidelity loss~\cite{chan2021understanding}.}
    \vspace{-0.3cm}
    \label{tab:alignment}
    \begin{center}
        \tabcolsep=0.15cm
        \scalebox{0.95}{
            \begin{tabular}{c|c|c||c}
                \hline
                          & w/o Flow & Offset-Fidelity Loss~\cite{chan2021understanding} & \textbf{Ours} \\\hline
                PSNR (dB) & 27.44    & 30.22                                             & 32.39         \\\hline
            \end{tabular}}
    \end{center}
    \vspace{-0.5cm}
\end{table}
\begin{figure}[t]
    \begin{center}
        \includegraphics[width=0.49\textwidth]{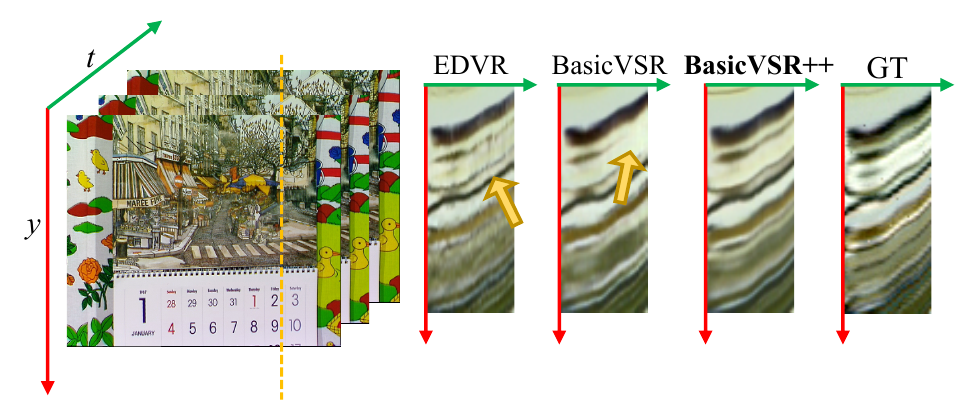}
        \vspace{-0.7cm}
        \caption{\textbf{Comparison of temporal profile.} We select a column (orange dotted lines) and observe the changes across time. The profile from EDVR possesses noise, indicating flickering artifacts. The profile from BasicVSR still contains discontinuity. By better aggregating the long-term information, the profile from BasicVSR++ demonstrates a smoother transition.}
        \label{fig:profile}
    \end{center}
    \vspace{-0.7cm}
\end{figure}

\noindent\textbf{Temporal Consistency}.
Here, we examine the temporal consistency, which is another important direction in VSR.
The recurrent framework intrinsically maintains a better temporal consistency in comparison to the sliding-window framework.
In the sliding-window framework (\eg,~EDVR~\cite{wang2019edvr}), each frame is reconstructed independently. In such a design, the consistency between the outputs cannot be guaranteed.
In contrast, in the recurrent framework (\eg,~BasicVSR~\cite{chan2021basicvsr}), the outputs are related through the propagation of the intermediate features. The temporal propagation essentially helps maintaining better temporal consistency.

In Fig.~\ref{fig:profile} we show a comparison of the temporal profiles between BasicVSR++ and two state-of-the-art methods -- EDVR and BasicVSR.
For the sliding-window method, the temporal profile from EDVR contains significant noise, indicating flickering artifacts in the output video.
In contrast, for recurrent networks, without explicit modeling of temporal consistency, the profiles from BasicVSR and BasicVSR++ demonstrate better consistencies.
However, the profile from BasicVSR still contains discontinuity. Benefit from our enhanced propagation and alignment, BasicVSR++ is able to aggregate richer information from the video frames, showing smoother temporal transition.
The video results are given in the supplementary material.

\section{NTIRE 2021 Challenge Results}
\begin{figure}[t]
    \begin{center}
        \includegraphics[width=0.49\textwidth]{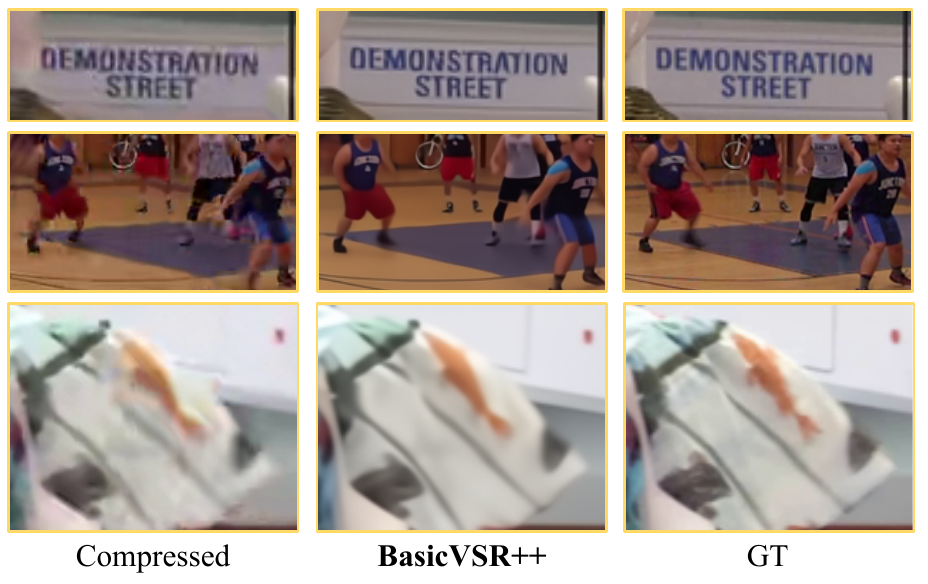}
        \caption{\textbf{Results on compressed video enhancement.} The outputs clearly possesses fewer artifacts, and the details are shown more clearly.}
        \label{fig:uncompress}
    \end{center}
    \vspace{-0.7cm}
\end{figure}

In NTIRE 2021, BasicVSR++ wins the video super-resolution track~\cite{son2021ntire} with a compact and efficient structure. In addition to VSR, BasicVSR++ generalizes well to other restoration tasks. BasicVSR++ obtains two champions and one runner-up in the compressed video enhancement challenge~\cite{yang2021ntire}. Fig.~\ref{fig:uncompress} shows the restoration results of three different patches of compressed videos. BasicVSR++ successfully reduces the artifacts and produces outputs with much better qualities. The promising performance in the competitions demonstrate the generalizability and versatility of BasicVSR++.

\section{Conclusion}
In this work, we redesign BasicVSR with two novel components to enhance its propagation and alignment performance for the task of video super-resolution. Our model BasicVSR++ outperforms existing state of the arts by a large margin while maintaining efficiency. These designs generalizes well to other video restoration tasks including compressed video enhancement. These components are generic and we speculate that they will be useful for other video-based enhancement or restoration tasks such as deblurring and denoising.

{\small
    \bibliographystyle{ieee_fullname}
    \bibliography{short,citations}
}
\appendix
\section{Network Architecture}
We use pretrained SPyNet~\cite{ranjan2017optical} as our flow network. The number of residual blocks for the initial feature extraction is set to $5$, and the number of residual blocks for each propagation branch is set to $7$. The feature channel is set to $64$.

The architecture of our second-order deformable alignment is highly similar to the first-order counterpart (Fig. 3 in the main paper). The only difference is that the pre-aligned features and optical flows from different timesteps are concatenated, and passed to the offset estimation module $\mathcal{C}^o$ and mask estimation module $\mathcal{C}^m$. Their architectures are detailed in Table~\ref{tab:offset}.
We set the DCN kernel size to $3$ and the number of deformable groups to $16$. Codes will be released.

\section{Experimental Settings}
\noindent\textbf{Datasets.}
Two widely-used datasets are adopted for training: REDS~\cite{nah2019ntire} and Vimeo-90K~\cite{xue2019video}.
For REDS, following BasicVSR~\cite{chan2021basicvsr}, we use REDS4\footnote{Clips 000, 011, 015, 020 of REDS training set.} as our test set and REDSval4\footnote{Clips 000, 001, 006, 017 of REDS validation set.} as our validation set. The remaining clips are used for training.
We use Vid4~\cite{liu2014bayesian}, UDM10~\cite{yi2019progressive}, and Vimeo-90K-T~\cite{xue2019video} as test sets along with Vimeo-90K.

\noindent\textbf{Degradations.}
All models are tested with $4{\times}$ downsampling using two degradations -- Bicubic (BI) and Blur Downsampling (BD). For BI, the MATLAB function \texttt{imresize} is used for downsampling. For BD, we blur the ground-truth by a Gaussian filter with $\sigma{=}1.6$, followed by a subsampling every four pixels.

\noindent\textbf{Training Settings.}
We adopt Adam optimizer~\cite{kingma2014adam} and Cosine Annealing scheme~\cite{loshchilov2016sgdr}. When trained on REDS, the initial learning rate of the main network and the flow network are set to $1{\times}10^{-4}$ and $2.5{\times}10^{-5}$, respectively. The total number of iterations is 600K, and the weights of the flow network are fixed during the first 5,000 iterations. The batch size is 8 and the patch size of input LR frames is $64{\times}64$. We use Charbonnier loss~\cite{charbonnier1994two} since it better handles outliers and improves the performance over the conventional $\ell_2$-loss~\cite{lai2017deep}. During training, $30$ LR frames are used as inputs.
Since Vimeo-90K contains only seven frames per sequence, networks trained solely on Vimeo-90K may not be able to capture long-term dependencies. Therefore, we initialize the model using the weights trained on REDS when trained on Vimeo-90K. The number of finetune iterations is 300K.

\noindent\textbf{Test Settings.}
We take the full video sequence as inputs to explore information from all video frames for restoration.

\begin{table}[!t]
    \caption{\textbf{Architectures of $\mathcal{C}^o$ and $\mathcal{C}^m$}. The two modules share the first six layers. They can be implemented as a stack of convolutions followed by a channel-splitting. The arguments in the convolution layer are \textit{input channels}, \textit{output channels}, and \textit{kernel size}, respectively.}
    \label{tab:offset}
    \vspace{-1cm}
    \begin{center}
        \tabcolsep=0.15cm
        \vspace{0.5cm}
        \scalebox{0.9}{
            \begin{tabular}{|c|c|c|}
                \hline
                Layer & $\mathcal{C}^o$                       & $\mathcal{C}^m$  \\\hline
                1.    & \multicolumn{2}{c|}{conv(196, 64, 3)}                    \\\hline
                2.    & \multicolumn{2}{c|}{LeakyReLU(0.1)}                      \\\hline
                3.    & \multicolumn{2}{c|}{conv(64, 64, 3)}                     \\\hline
                4.    & \multicolumn{2}{c|}{LeakyReLU(0.1)}                      \\\hline
                5.    & \multicolumn{2}{c|}{conv(64, 64, 3)}                     \\\hline
                6.    & \multicolumn{2}{c|}{LeakyReLU(0.1)}                      \\\hline
                7.    & conv(64, 288, 3)                      & conv(64, 144, 3) \\\hline
            \end{tabular}}
        \vspace{-0.4cm}
    \end{center}
\end{table}
\section{Qualitative Comparisons}
In this section, we provide additional qualitative comparisons on REDS4~\cite{nah2019ntire}, UDM10~\cite{yi2019progressive}, Vimeo-90K~\cite{xue2019video}, and Vid4~\cite{liu2014bayesian}. From the examples, we see that BasicVSR++ is able to restore the fine details, leading to plausible results. A video demo is also provided in the submitted zip file.

\begin{figure*}[!t]
    \begin{center}
        \includegraphics[width=0.99\textwidth]{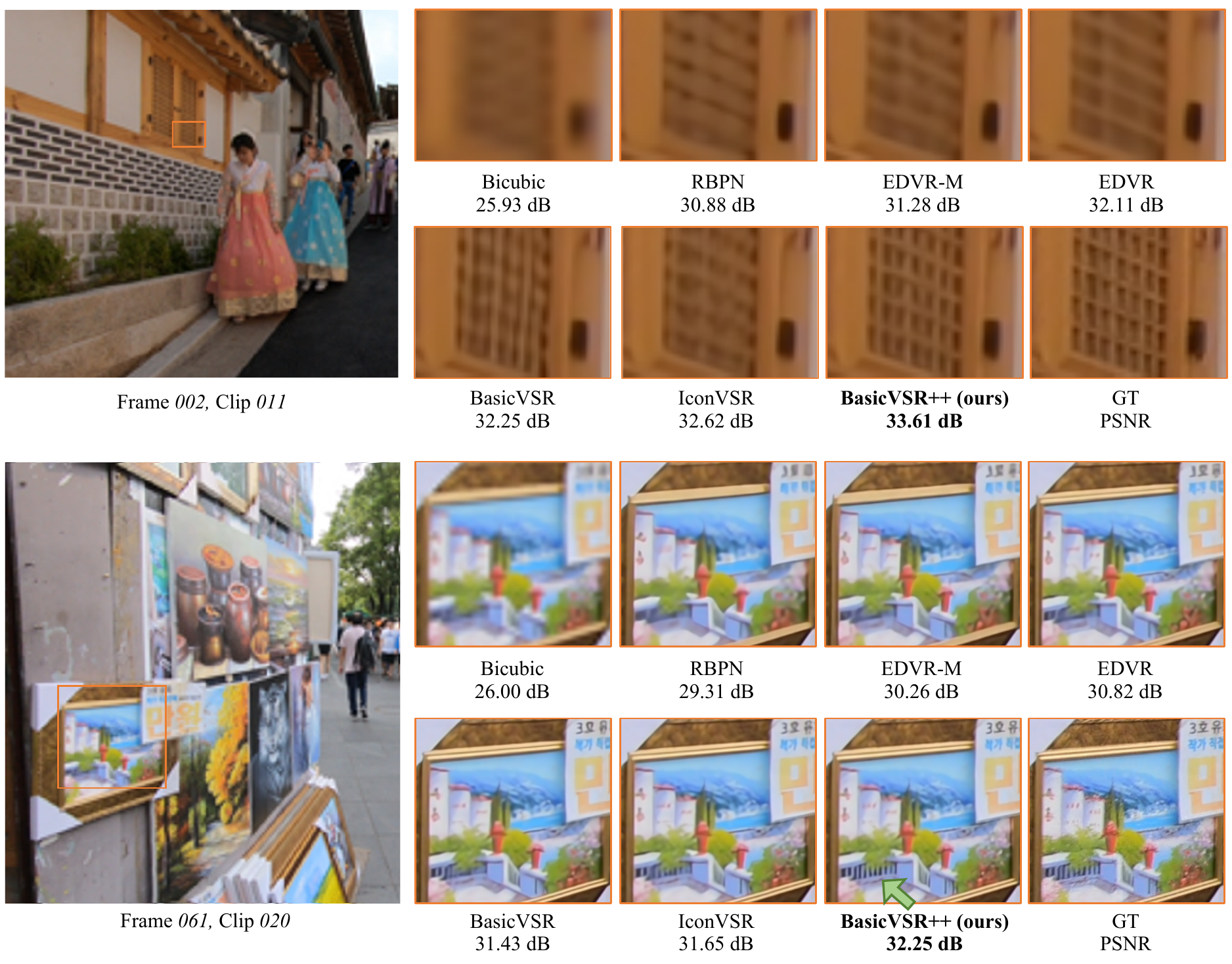}
        \vskip -0.35cm
        \caption{Qualitative comparison on REDS4~\cite{wang2019edvr}.}
        \label{fig:reds4}
    \end{center}
    \vspace{-0.2cm}
\end{figure*}
\begin{figure*}[!t]
    \begin{center}
        \includegraphics[width=0.99\textwidth]{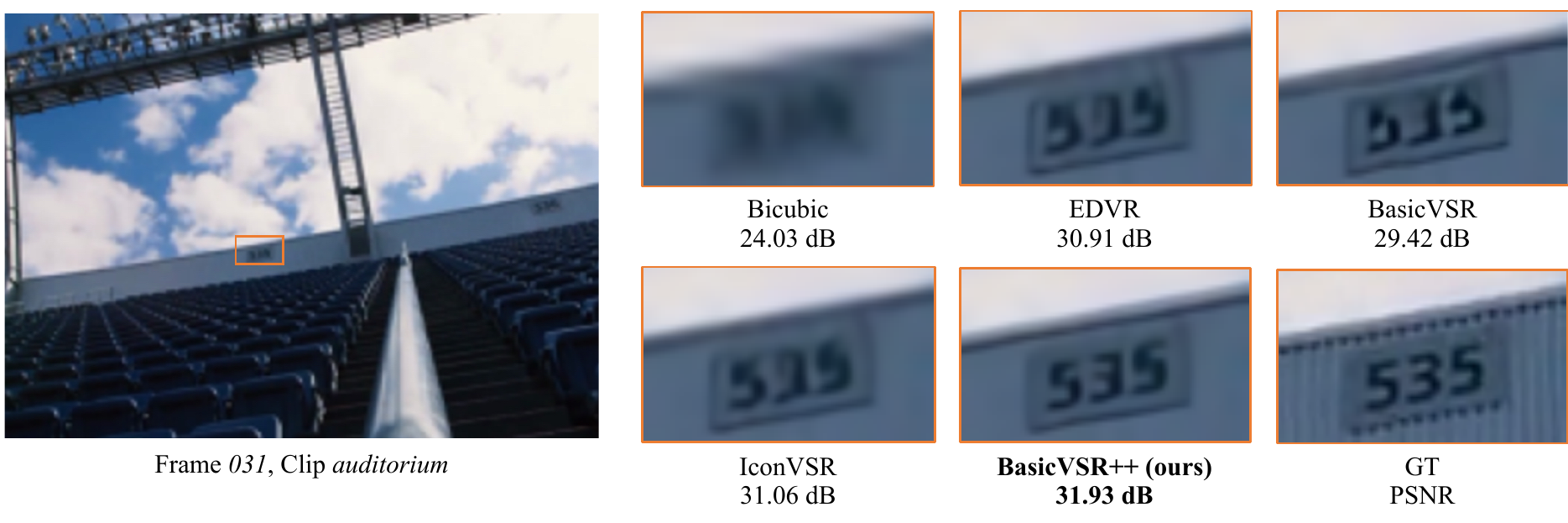}
        \vskip -0.35cm
        \caption{Qualitative comparison on UDM10~\cite{yi2019progressive}.}
        \label{fig:udm10}
    \end{center}
    \vspace{-0.2cm}
\end{figure*}
\begin{figure*}[!t]
    \begin{center}
        \includegraphics[width=0.99\textwidth]{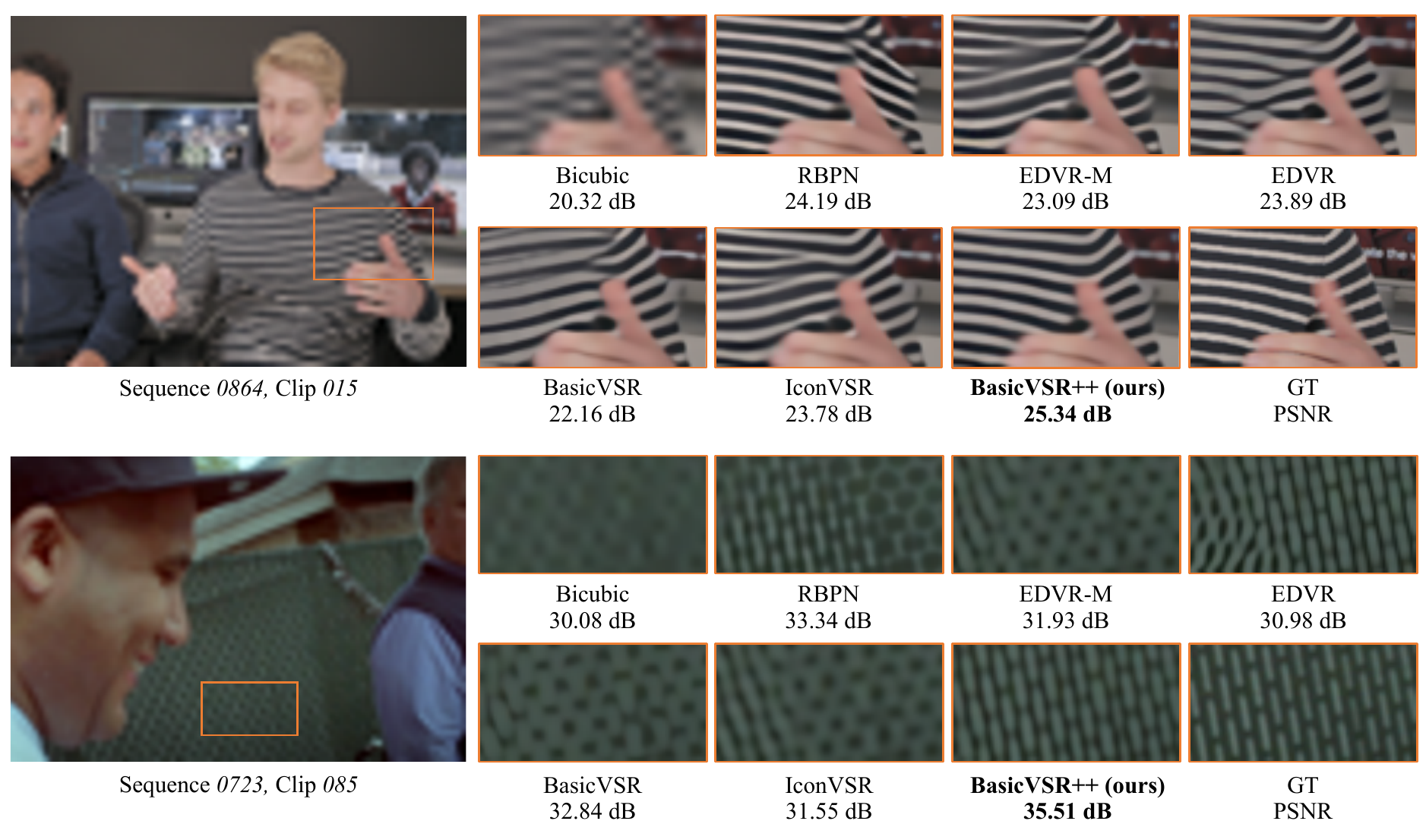}
        \vskip -0.35cm
        \caption{Qualitative comparison on Vimeo-90K-T~\cite{xue2019video}.}
        \label{fig:vimeo}
    \end{center}
    \vspace{-0.2cm}
\end{figure*}
\begin{figure*}[!t]
    \begin{center}
        \includegraphics[width=0.99\textwidth]{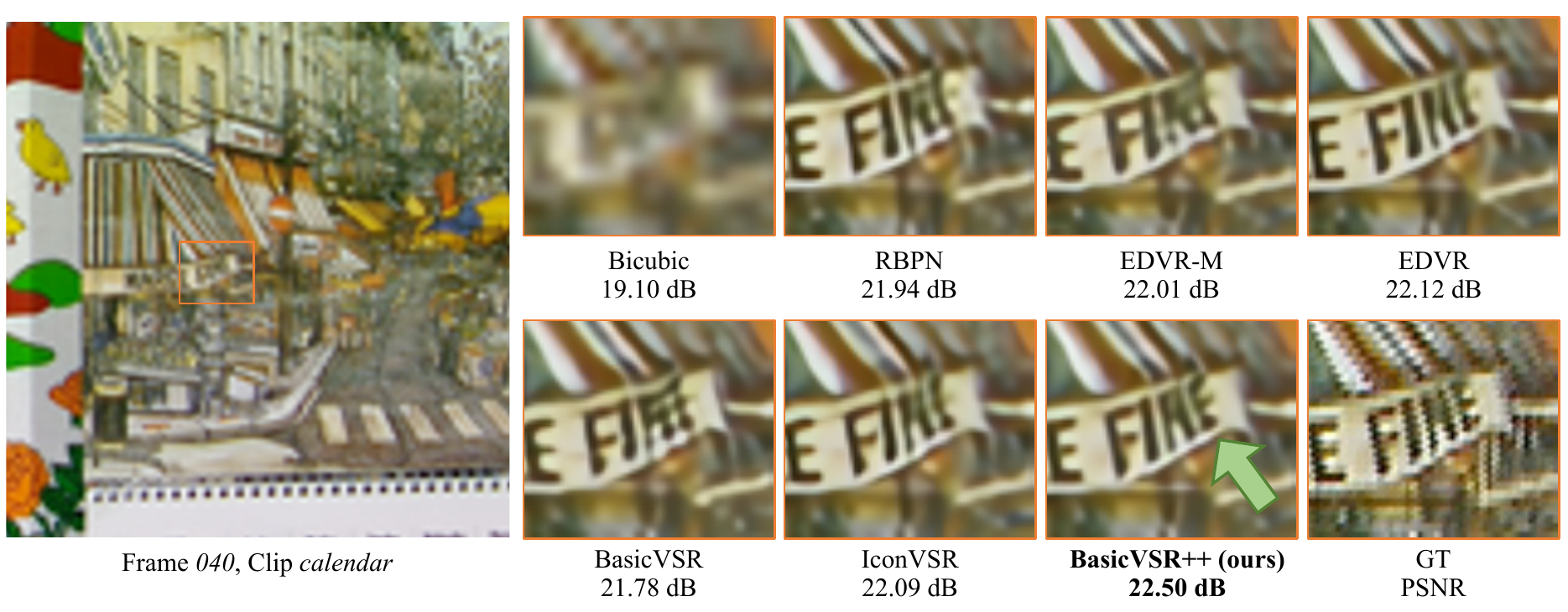}
        \vskip -0.35cm
        \caption{Qualitative comparison on Vid4~\cite{liu2014bayesian}.}
        \label{fig:vid4}
    \end{center}
    \vspace{-0.2cm}
\end{figure*}

\end{document}